\documentclass[sigconf]{acmart}

\AtBeginDocument{%
  }

\usepackage{multirow,makecell,booktabs}
\usepackage{float,subfig,color}
\usepackage{enumitem}
\usepackage{algorithmic,algorithm}
\usepackage{comment}
\usepackage{bm}
\usepackage{balance}

\usepackage{amsmath,amsfonts,bm}

\def\eqref#1{eq.~(\ref{#1})}
\def\Eqref#1{Eq.~(\ref{#1})}

\def\1{\bm{1}}

\def\rd{{\textnormal{d}}}

\def\vx{{\bm{x}}}

\def\vz{{\bm{z}}}

\DeclareMathAlphabet{\mathsfit}{\encodingdefault}{\sfdefault}{m}{sl}
\SetMathAlphabet{\mathsfit}{bold}{\encodingdefault}{\sfdefault}{bx}{n}

\usepackage{array}
\newcommand{\PreserveBackslash}[1]{\let\temp=\\#1\let\\=\temp}
\newcolumntype{C}[1]{>{\PreserveBackslash\centering}m{#1}}
\newcolumntype{R}[1]{>{\PreserveBackslash\raggedleft}m{#1}}
\newcolumntype{L}[1]{>{\PreserveBackslash\raggedright}m{#1}}
\newcolumntype{P}[1]{>{\raggedright\arraybackslash}m{#1}}

\usepackage{soul}
\let\ul\underline

\newcommand{\std}[1]{\tiny{$\pm$#1}}
\newcommand{\avgstd}[1]{\tiny{\;(#1)}}

\usepackage{xcolor}

\copyrightyear{2025}
\acmYear{2025}
\setcopyright{acmlicensed}
\acmConference[CIKM '25] {Proceedings of the 34th ACM International Conference on Information and Knowledge Management}{ November 10--14, 2025}{Seoul, Republic of Korea.}
\acmBooktitle{Proceedings of the 34th ACM International Conference on Information and Knowledge Management (CIKM '25), November 10--14, 2025, Seoul, Republic of Korea}
\acmISBN{979-8-4007-2040-6/2025/11}
\acmDOI{10.1145/3746252.3760996}
\begin{document}

\title{\texttt{TANDEM}: Temporal Attention-guided Neural Differential Equations for Missingness in Time Series Classification}

\author{YongKyung Oh}
\email{yongkyungoh@mednet.ucla.edu}
\affiliation{%
  \institution{University of California, Los Angeles,}
  \city{Los Angeles}
  \state{CA}
  \country{USA}
}

\author{Dong-Young Lim}
\email{dlim@unist.ac.kr}
\affiliation{%
  \institution{Ulsan National Institute of Science and Technology,}
  \city{Ulsan}
  \country{Republic of Korea}
}

\author{Sungil Kim\footnotemark[1]}
\email{sungil.kim@unist.ac.kr}
\affiliation{%
  \institution{Ulsan National Institute of Science and Technology,}
  \city{Ulsan}
  \country{Republic of Korea}
}

\author{Alex A. T. Bui}
\email{buia@mii.ucla.edu}
\affiliation{%
  \institution{University of California, Los Angeles,}
  \city{Los Angeles}
  \state{CA}
  \country{USA}
}

\authornote{Corresponding authors}

\renewcommand{\shortauthors}{Oh et al.}

\begin{abstract}
    Handling missing data in time series classification remains a significant challenge in various domains. Traditional methods often rely on imputation, which may introduce bias or fail to capture the underlying temporal dynamics. In this paper, we propose \texttt{TANDEM} (\emph{Temporal Attention-guided Neural Differential Equations for Missingness}), an attention-guided neural differential equation framework that effectively classifies time series data with missing values. Our approach integrates raw observation, interpolated control path, and continuous latent dynamics through a novel attention mechanism, allowing the model to focus on the most informative aspects of the data. We evaluate \texttt{TANDEM} on 30 benchmark datasets and a real-world medical dataset, demonstrating its superiority over existing state-of-the-art methods. Our framework not only improves classification accuracy but also provides insights into the handling of missing data, making it a valuable tool in practice.
\end{abstract}




\maketitle

\section{Introduction}
Time series data are fundamental in many domains but are often suffer from missing values due to sensor issues, data corruption, or irregular sampling \citep{rubin_inference_1976, little_statistical_2019}. Traditional methods for handling missingness, such as deletion or simple imputation, can distort temporal dynamics and introduce bias, especially with non-random or extensive missing data \citep{janssen_missing_2010, emmanuel_survey_2021}. While generative imputation models \citep{cao_brits_2018, yoon_gain_2018} offer improvements, they often treat imputation as a separate preprocessing step, potentially neglecting uncertainties and complex dependencies in irregularly sampled time series.

Neural Differential Equations (NDEs), including Neural ODEs \citep{chen_neural_2018}, CDEs \citep{kidger_neural_2020}, and SDEs \citep{tzen_neural_2019, oh_stable_2024}, provide a robust framework for modeling continuous-time dynamics, adept at handling complex and irregularly sampled data. Neural CDEs, for example, effectively model irregular series by using an input-derived control path to drive the hidden state dynamics. However, NDEs alone do not explicitly address how to best utilize available information when observations are sparse or incomplete for classification tasks \citep{oh_comprehensive_2025}. Attention mechanisms \citep{vaswani_attention_2017}, renowned for enabling models to focus on salient input segments, offer a promising way to enhance NDEs in such scenarios by adaptively weighing data importance \citep{shukla_multi-time_2021, y_lee_multi-view_2022}.

While the integration of NDEs and attention is an active research area~\citep{jhin_ace-node_2021, j_-t_chien_continuous-time_2021, chen_contiformer_2023, jhin_attentive_2024, tong_neural_2025}, a comprehensive framework that adaptively fuses diverse NDE-derived representations under varying missingness conditions, across multiple NDE backbones, remains less explored.
To address this, we propose \texttt{TANDEM} (\emph{Temporal Attention-guided Neural Differential Equations for Missingness}). \texttt{TANDEM} synergistically combines NDEs with a novel attention and gating mechanism to classify time series with missing values. 
It integrates three feature streams: raw observations, control path, and continuous NDE-derived latent dynamics. A Gumbel-Sigmoid gating module with multi-head attention fuses these streams and prioritizes informative signals under missingness.
Our contributions are:
\begin{itemize}
    \item Novel framework, \texttt{TANDEM}, that integrates raw, interpolated, and latent NDE-based representations to handle missingness in time series. 
    Flexible Gumbel-Sigmoid gating mechanism with multi-head attention for adaptive fusion of these feature streams, enhancing robustness under high missingness.
    \item A modular design supporting various NDE backbones (Neural ODEs, CDEs, SDEs), allowing users to tailor the continuous-time model to their data.
    \item Extensive validation on 30 UCR/UEA benchmarks and a real-world medical dataset, demonstrating superior classification accuracy and robustness over state-of-the-art methods across diverse missingness levels.
\end{itemize}

This paper is organized as follows: Section~\ref{sec:Related_works} reviews related work. Section~\ref{sec:Methodology} details the \texttt{TANDEM} framework. Experimental results are presented in Section~\ref{sec:Experiment} (benchmarks) and Section~\ref{sec:Real_world} (real-world application). Section~\ref{sec:Conclusion} concludes and discusses future work.

\begin{figure*}[!htbp]
    \centering\captionsetup{skip=5pt}
    \includegraphics[width=0.95\linewidth]{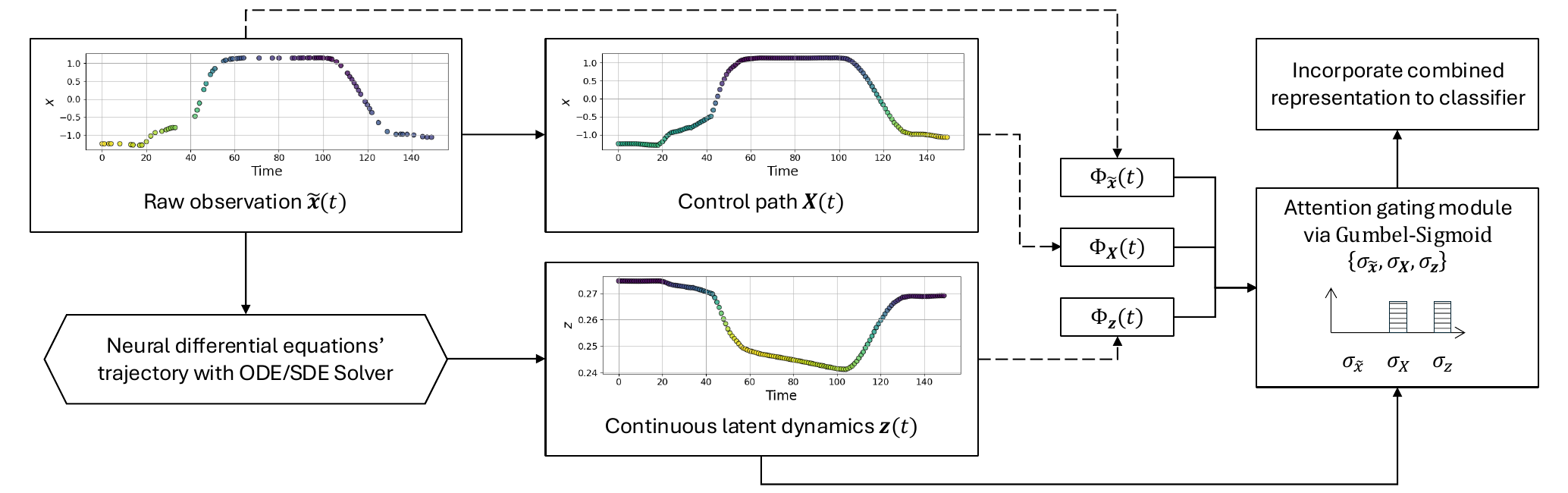}
    \caption{Conceptual overview of the \texttt{TANDEM} framework. For a given time series with potentially missing values, three distinct feature streams are processed: (i) the raw observation $\tilde{\vx}(t)$, (ii) an interpolated, piecewise-smooth control path $\bm{X}(t)$, and (iii) continuous latent dynamics $\vz(t)$ derived from an NDE backbone. Each stream is individually refined by attention mechanisms, resulting in attended representations $\Phi_{\tilde{\vx}}(t)$, $\Phi_{\bm{X}}(t)$, and $\Phi_{\vz}(t)$. Colors represent the learned temporal attention scores for each stream.
    These attended representations are then adaptively weighted by learnable Gumbel-Sigmoid gates ($\sigma_{\tilde{\vx}}$, $\sigma_{\bm{X}}$, $\sigma_{\vz}$) which determine the contribution of each stream. The resulting fused representation, $\bar{\mathcal{Z}}(t)$, is subsequently passed to a classifier.
    }
    \label{fig:overview}
    \Description{}
    \vspace{-1em}
\end{figure*}

\section{Related Works}\label{sec:Related_works}
NDEs have emerged as a powerful paradigm for modeling continuous-time dynamics \citep{oh_comprehensive_2025}.
The foundational Neural ODEs \citep{chen_neural_2018} parameterize the derivative of a system's hidden state with a neural network, offering flexibility for irregularly sampled data.
Enhancements include Augmented Neural ODEs \citep{dupont_augmented_2019} for increased expressiveness, and Latent ODEs \citep{rubanova_latent_2019} which learn dynamics in a latent space, beneficial for series with missingness.
Neural CDEs \citep{kidger_neural_2020} generalize Neural ODEs by incorporating an input-derived control path, making them highly effective for irregular time series. This was further extended by Neural Rough Differential Equations (Neural RDEs) using rough path theory to handle noisy, long sequences \citep{morrill_neural_2021}.
Concurrently, Neural SDEs have been developed to capture stochastic dynamics. \citet{tzen_neural_2019} introduced Neural SDEs for learning stochastic processes, \citet{li_scalable_2020} addressed scalable gradient computation for them, and \citet{oh_stable_2024} analyzed their stability for irregular data.
While these NDE methods excel at modeling continuous dynamics, they often lack explicit mechanisms to adaptively prioritize information from different feature representations when faced with missing observations, a gap our work addresses.

\paragraph{Attention Mechanisms in Sequence Modeling.}
Attention mechanisms, first prominent in neural machine translation \citep{bahdanau_neural_2015}, enable models to dynamically focus on relevant parts of an input sequence. The transformer architecture \citep{vaswani_attention_2017}, centered around self-attention, further established this approach for capturing dependencies in sequences.
In time series analysis, attention has been used to capture temporal patterns and feature importance. For instance, \citet{qin_dual-stage_2017} proposed dual-stage attention for prediction, \citet{shih_temporal_2019} introduced temporal pattern attention for multivariate forecasting, and \citet{lim_temporal_2021} developed temporal fusion transformers for interpretable multi-horizon forecasting. These methods highlight the utility of attention in focusing on critical time steps or features.

\paragraph{Integrating NDEs and Attention Mechanisms.}
The integration of NDEs with attention mechanisms, which combines continuous-time modeling with dynamic input weighting, is an active research area. Initial explorations often occurred in Natural Language Processing (NLP). 
ODE Transformer \citep{li_ode_2022} models attention dynamics as a discretized ODE process for sequence generation tasks. \citet{zhang_continuous_2021} introduced a continuous self-attention mechanism via Neural ODEs, evolving representations over continuous depth. More recently, \citet{tong_neural_2025} further analyzed the internal dynamics of Neural ODE Transformers and proposed adaptive fine-tuning. 

In the time series domain, efforts to combine these paradigms include ACE-NODE (Attentive Co-Evolving Neural Ordinary Differential Equations) by \citet{jhin_ace-node_2021}, which integrates attention with Neural ODEs, and research by \citet{j_-t_chien_continuous-time_2021} on continuous-time self-attention within NDEs. ANCDE (Attentive Neural Controlled Differential Equation)~\citep{jhin_attentive_2024} applied attention mechanisms specifically to the control paths in Neural CDEs.

While these represent valuable advancements, they often target specific NDE backbones or particular attention integration strategies. A comprehensive framework that supports multiple NDE types (Neural ODE, CDE, SDE), adaptively fuses diverse feature representations via explicit gating, and is tailored for robust classification under various missing data conditions, remains a key area for development. \texttt{TANDEM} is designed to address this gap by providing a unified, attention-guided mechanism to dynamically weigh and combine these distinct information sources.

\paragraph{Differentiable Gating for Feature Selection}
{Relying on a single stream can obscure missingness patterns or limit expressivity \citep{lipton_directly_2016}. Prior works such as \citep{shukla_multi-time_2021, y_lee_multi-view_2022} use attention to learn from raw input features directly, but often lack mechanisms to selectively weight fundamentally different representation types. 
On the other hand, interpolated paths provide global trends \citep{kidger_neural_2020}, and NDEs capture latent dynamics \citep{rubanova_latent_2019}. Although these features are complementary, there remains a gap in effectively integrating them.}
Effectively selecting relevant information streams is a core challenge in multi-source integration. The Gumbel-Softmax trick \citep{jang_categorical_2017, maddison_concrete_2017} addresses this by enabling differentiable, near-discrete selection through continuous relaxation of categorical variables, allowing end-to-end training of selection mechanisms within model.
Based on that, \texttt{TANDEM} tackles this challenge by integrating raw, interpolated, and latent features using Gumbel-based differentiable gating, enabling adaptive and near-discrete fusion guided by multi-view learning principles.

\section{Methodology}\label{sec:Methodology}

The \texttt{TANDEM} framework is designed to perform robust time series classification. 
It achieves this by adaptively integrating multiple feature representations derived from the input series and its continuous latent dynamics, as conceptually illustrated in Figure~\ref{fig:overview}. 

\paragraph{Problem Formulation.}
We consider an $\mathbb{R}^d$-valued time series as $\vx = (x(1), \ldots, x(T))$, where each observation $x(t)$ is a $d$-dimensional vector. In many practical settings, observations can be missing or incomplete. We represent this missingness using a binary mask $M = (M(1), \ldots, M(T))$, where $M(t) \in \{0, 1\}^d$. An entry $M(t)_j = 0$ signifies that the $j$-th dimension of $x(t)$ is missing, while $M(t)_j = 1$ indicates it is observed. The resulting time series with potentially missing values is denoted as $\tilde{\vx} = (\tilde{x}(1), \ldots, \tilde{x}(T))$, where $\tilde{x}(t) = M(t) \odot x(t)$, with $\odot$ representing the element-wise product. 
Given a dataset $\mathcal{D} = \{(\tilde{\vx}^{(i)}, M^{(i)}, y^{(i)})\}_{i=1}^N$ consisting of $N$ such time series, where $y^{(i)} \in \{1, \ldots, C\}$ is the class label for the $i$-th series, our objective is to learn a classifier $f: \mathbb{R}^{d \times T} \times \{0, 1\}^{d \times T} \rightarrow \{1, \ldots, C\}$ that accurately predicts $y$ for a given $\tilde{\vx}$ and $M$.

\subsection{Multi-Perspective Feature Extraction}\label{sec:feature_extraction}
To effectively handle missing data and capture the underlying temporal dynamics, \texttt{TANDEM} utilizes three distinct yet complementary feature representations extracted from the input time series. As illustrated in Figure~\ref{fig:overview}, these features offer different perspectives on the data: (i) the raw, potentially incomplete, observations; (ii) an interpolated continuous control path; and (iii) continuous latent dynamics modeled by a NDE. 
These representations are complementary to one another and are subsequently fused.

\paragraph{Raw Observation ($\tilde{\vx}(t)$).}
The primary input is the raw observed time series $\tilde{\vx}(t) \in \mathbb{R}^{d}$. This stream provides the model with the actual observed data points at discrete time steps, including any missing entries.  
Preserving raw observations is crucial because the pattern of missingness itself can sometimes be informative \citep{lipton_directly_2016}, and the observed values, however sparse, constitute the most direct evidence of the system's state.

\paragraph{Interpolated Control Path ($\bm{X}(t)$).}
To derive a continuous representation from potentially irregular and incomplete raw observations, we construct an interpolated control path, $\bm{X}(t)$. This path offers a smooth, continuous trajectory that spans missing segments and effectively regularizes the input signal. 
This approach is chosen for its ability to produce a smooth curve that passes through all observed data points, providing a continuously differentiable path that is beneficial for certain NDEs, such as Neural CDEs \citep{kidger_neural_2020}. Given the set of observed points $\{ (t_k, \tilde{x}(t_k)) \}$ where $M(t_k)_j = 1$ for relevant dimensions, the spline $\bm{X}(s)$ for $s \in [0, T]$ yields this continuous function. 
The control path $\bm{X}(t)$ acts as an initial continuous approximation that guides the NDE backbone and provides a complete representation of the time series trajectory for subsequent attention-based processing.

\paragraph{Continuous Latent Dynamics ($\vz(t)$).}
To model the underlying generative process and learn a representation of the system's dynamics in a continuous latent space, we utilize NDEs as backbone models. The hidden trajectory $\vz(t) \in \mathbb{R}^{d_z}$ aims to capture the evolving state of the system. Based on the assumed nature of these dynamics, different NDE backbones can be employed within our framework:
\begin{itemize}
    \item \textbf{Neural ODEs} \citep{chen_neural_2018} describe deterministic dynamics:
        \begin{align}\label{eq:neural_ode}
        \vz(t) = \vz(0) + \int_0^t f(s, \vz(s); \theta_f) \; \rd s.
        \end{align}
        Here, $f$ is a neural network parameterizing the derivative, and the initial state $\vz(0)$ is typically learned via another neural network $\eta(\vx_{\text{init}}; \theta_\eta)$ based on an initial segment or summary of $\tilde{\vx}$.
    \item \textbf{Neural CDEs} \citep{kidger_neural_2020} model dynamics as controlled by the input path $\bm{X}(t)$:
        \begin{align}\label{eq:neural_cde}
        \vz(t) = \vz(0) + \int_0^t f(s, \vz(s); \theta_f) \; \rd \bm{X}(s).
        \end{align}
        The Riemann-Stieltjes integral allows $\vz(t)$ to respond dynamically to the continuous signal $\bm{X}(t)$.
    \item \textbf{Neural SDEs} \citep{tzen_neural_2019, oh_stable_2024} introduce stochasticity:
        \begin{align}\label{eq:neural_sde}
        \vz(t) = \vz(0) + \int_0^t f(s, \vz(s); \theta_f) \; \rd s + \int_0^t g(s,\vz(s);\theta_g) \;  \rd B(s),
        \end{align}
        where $f$ is the drift function, $g$ is the diffusion function, and $B(t)$ represents a Brownian motion.
\end{itemize}
These NDEs yield a continuous representation $\vz(t)$ that naturally accommodates irregular sampling. This representation provides a dynamically evolving summary of the time series, capable of inferring behavior even across missing segments. The flexibility to choose among different NDEs allows for tailoring the model to various types of temporal dependencies. The vector fields $f$ (and $g$) are parameterized by neural networks.

\subsection{Adaptive Feature Fusion with Gating}\label{sec:adaptive_fusion}
To effectively leverage the complementary information from the three extracted feature streams, we introduce an adaptive fusion mechanism. This mechanism comprises two key components: (1) a feature-wise multi-head attention module to capture the relevance of different dimensions within each feature stream at each time step, and (2) a stream-wise gating module employing a Gumbel-Sigmoid distribution to dynamically select or weigh the contribution of each entire feature stream. 

\subsubsection{Feature-wise Multi-head Attention}\label{sec:attention_module}
We first apply a multi-head attention mechanism \citep{vaswani_attention_2017} independently to each of the three feature streams: $\tilde{\vx}(t) \in \mathbb{R}^{d}$, $\bm{X}(t) \in \mathbb{R}^{d}$, and $\vz(t) \in \mathbb{R}^{d_z}$. This step allows the model to weigh the importance of different dimensions or learned features within each individual stream at every time point $t$.
For a generic feature stream $\phi_k(t)$ (where $k$ indexes one of the three streams: raw, control, or latent), we project it into query ($Q_k^h(t)$), key ($K_k^h(t)$), and value ($V_k^h(t)$) representations for each attention head $h \in \{1, \ldots, H\}$:
\begin{align}
    Q_k^h(t) = W_{Q,k}^h \phi_k(t), \quad K_k^h(t) = W_{K,k}^h \phi_k(t), \quad V_k^h(t) = W_{V,k}^h \phi_k(t),
\end{align}
where $W_{Q,k}^h, W_{K,k}^h, W_{V,k}^h$ are trainable weight matrices specific to feature stream $k$ and head $h$. The attention weights for each head are computed using scaled dot-product attention:
\begin{align}
    \mathrm{attn\_weights}_k^h(t) = \text{softmax}\left(\frac{Q_k^h(t)^T K_k^h(t)}{\sqrt{d_{key}}}\right),
\end{align}
where $d_{key}$ is the dimension of the key vectors. The attended feature representation for stream $k$ from head $h$ is then $\Phi_k^h(t) = \mathrm{attn\_weights}_k^h(t) V_k^h(t)$.
The outputs from all heads are subsequently concatenated and linearly projected to produce the final attended representation for each stream:
\begin{align}
    \Phi_k(t) = W_{O,k} [\Phi_k^1(t); \ldots; \Phi_k^H(t)],
\end{align}
where $W_{O,k}$ is a trainable output weight matrix for stream $k$. This process yields the attended representations $\Phi_{\tilde{\vx}}(t)$, $\Phi_{\bm{X}}(t)$, and $\Phi_{\vz}(t)$.
In Figure~\ref{fig:overview}, the learned attention scores are visualized using different colors. For the latent dynamics $\vz$, we plot the average value of the hidden states for clarity.
 
An element-wise sigmoid function is applied to each $\Phi_k(t)$. The primary role of this sigmoid activation is to normalize the output of the attention module for each stream. 
It aims to ensure that no single attended stream disproportionately influences the fusion due to its scale, rather than its learned importance, before the explicit gating mechanism determines its contribution.

\subsubsection{Stream-wise Gating via Gumbel-Sigmoid}\label{sec:gating_module}
While feature-wise attention refines each stream internally, a mechanism is also needed to determine the overall importance or reliability of each entire attended stream ($\Phi_{\tilde{\vx}}(t)$, $\Phi_{\bm{X}}(t)$, $\Phi_{\vz}(t)$), particularly under varying conditions of data missingness. For this purpose, we introduce a stream-wise gating mechanism employing a Gumbel-Sigmoid distribution. 
We opt for the Gumbel-Sigmoid \citep{jang_categorical_2017, maddison_concrete_2017} over a standard softmax across streams because it allows the model to select any subset of features independently. This contrasts with softmax, which forces a fixed probability mass (summing to 1) to be distributed across the streams.

We introduce learnable logits, $\ell_{\tilde{\vx}}$, $\ell_{\bm{X}}$, and $\ell_{\vz}$, which correspond to the desirability or relevance of each attended feature stream. The gating value  $\sigma_k$ for each stream $k \in \{\tilde{\vx}, \bm{X}, \vz\}$ is then computed using the Gumbel-Sigmoid function:
\begin{equation}
\label{eq:gumbel_sigmoid}
\sigma_k = \text{Sigmoid}\left( \frac{\ell_k + \mathcal{G}_k}{\tau} \right),
\end{equation}
where $\mathcal{G}_k = -\log(-\log(U_k))$ with $U_k \sim \text{Uniform}(0, 1)$ are independent Gumbel noise samples. The temperature parameter $\tau > 0$ controls the smoothness of this approximation: as $\tau \to 0$, $\sigma_k$ approaches a Bernoulli variable, thereby emulating a hard selection. During the training phase, we utilize a fixed temperature $\tau=1.0$ to ensure differentiability and enable gradient propagation. 
For inference, one can either use these soft gate values directly or apply hard thresholding for an explicit selection:
\begin{equation}
\label{eq:hard_selection}
\sigma_k^{\text{hard}} = \mathbb{I}(\sigma_k > 0.5),
\end{equation}
where $\mathbb{I}(\cdot)$ denotes the indicator function. In this study, we use the soft gates $\sigma_k$ derived from \Eqref{eq:gumbel_sigmoid} during both training and inference, unless specified otherwise in ablation studies (e.g., distinguishing `TANDEM (soft)` versus `TANDEM`).

The NDE's latent state $\vz(t)$ provides a foundational continuous-time dynamic representation. The information from the gated, attended feature streams is then combined to form the final representation for classification. 
The final combined feature representation $\bar{\mathcal{Z}}(t)$, which is passed to the classifier, is formed by a weighted aggregation of these attended and gated features:
\begin{align}\label{eq:final_fusion}
\bar{\mathcal{Z}}(t) = \sigma_{\tilde{\vx}}\Phi_{\tilde{\vx}}(t) \oplus \sigma_{\bm{X}}\Phi_{\bm{X}}(t) \oplus \sigma_{\vz}\Phi_{\vz}(t)
\end{align}
where $\oplus$ indicates a concatenation, depending on the input requirements of $f_{\text{classifier}}$. 
The key aspect is that the gates $\sigma_k$ modulate the influence of each fully attended feature stream $\Phi_k(t)$.

\subsection{Optimization for Classification}\label{sec:optimization_classification} 
The final fused feature representation $\bar{\mathcal{Z}}(t)$, as defined in \Eqref{eq:final_fusion}, provides a comprehensive summary of the time series, conditioned on the observed data and learned dynamics. For classification, we typically use the representation at the final time point, $\bar{\mathcal{Z}}(T)$, or alternatively, a pooled representation across all time steps, as input to a classification head. This head is tasked with mapping the learned temporal representation to a probability distribution over the $C$ possible class labels.

In this work, we employ a standard feed-forward neural network as the classifier: specifically, a two-layer Multilayer Perceptron (MLP). This MLP utilizes a \texttt{ReLU} activation function in its hidden layer, followed by a \texttt{Softmax} activation function in the output layer to produce the final class probabilities. The dimensionality of the hidden layer is typically set to be consistent with the dimension of the input representation $\bar{\mathcal{Z}}(T)$.
Denoting the classifier as $f_{\text{classifier}}(\cdot; \theta_{\text{classifier}})$, where $\theta_{\text{classifier}}$ represents its trainable parameters, the predicted class probabilities $\hat{y}$ are obtained as:
\begin{align}
\hat{y} = f_{\text{classifier}}(\bar{\mathcal{Z}}(T); \theta_{\text{classifier}}).
\end{align}
A crucial aspect of our \texttt{TANDEM} framework is that all its components are trained end-to-end. 
The optimization is driven by minimizing the standard cross-entropy loss function, suitable for multi-class classification tasks:
\begin{align}
\mathcal{L}(\Theta) = -\frac{1}{N} \sum_{i=1}^N \sum_{c=1}^C y_c^{(i)} \log \hat{y}_c^{(i)},
\end{align}
where $\Theta$ encompasses all trainable parameters of the model, $N$ is the number of time series instances in the training set, and $y_c^{(i)}$ is the one-hot encoded true label for the $i$-th time series for class $c$.
We employ gradient-based optimization algorithms, such as Adam \citep{kingma_adam_2015}, to minimize this loss function. Specific details regarding the hyperparameter settings, including learning rates and the use of potentially different learning rates for the final classifier layer to facilitate fine-tuning, are provided in Section~\ref{sec:Experiment} and the Appendix. This end-to-end training paradigm ensures that all parts of the model, from feature extraction and NDE modeling to attention-based fusion and gating, are optimized jointly to maximize classification performance on the target task, effectively learning to navigate the challenges posed by missing data.

\section{Experiment on benchmark datasets}\label{sec:Experiment}
All experiments were performed using a server on Ubuntu 22.04 LTS, equipped with an Intel(R) Xeon(R) Gold 6242 CPU and a cluster of NVIDIA A100 40GB GPUs. The source code for our experiments can be accessed at \url{https://github.com/yongkyung-oh/TANDEM}.

\begin{table}[htbp]
\scriptsize\centering\captionsetup{skip=5pt}
\caption{Data description for the time series datasets used in our experiments: dataset name, number of samples, number of classes, number of dimensions (variables) in each time series, and the length (time steps) of each time series.}\label{tab:data}
\begin{tabular}{@{}lrrrr@{}}
\toprule
\textbf{Dataset} & \textbf{$\#$ samples} & \textbf{$\#$ classes} & \textbf{$\#$ dimension} & \textbf{$\#$ length} \\ \midrule
\textbf{ArrowHead}                 & 211                           & 3                              & 1                            & 251                     \\
\textbf{Car}                       & 120                           & 4                              & 1                            & 577                     \\
\textbf{Coffee}                    & 56                            & 2                              & 1                            & 286                     \\
\textbf{GunPoint}                  & 200                           & 2                              & 1                            & 150                     \\
\textbf{Herring}                   & 128                           & 2                              & 1                            & 512                     \\
\textbf{Lightning2}                & 121                           & 2                              & 1                            & 637                     \\
\textbf{Lightning7}                & 143                           & 7                              & 1                            & 319                     \\
\textbf{Meat}                      & 120                           & 3                              & 1                            & 448                     \\
\textbf{OliveOil}                  & 60                            & 4                              & 1                            & 570                     \\
\textbf{Rock}                      & 70                            & 4                              & 1                            & 2844                    \\
\textbf{SmoothSubspace}            & 300                           & 3                              & 1                            & 15                      \\
\textbf{ToeSegmentation1}          & 268                           & 2                              & 1                            & 277                     \\
\textbf{ToeSegmentation2}          & 166                           & 2                              & 1                            & 343                     \\
\textbf{Trace}                     & 200                           & 4                              & 1                            & 275                     \\
\textbf{Wine}                      & 111                           & 2                              & 1                            & 234                     \\ \midrule
\textbf{ArticularyWordRecognition} & 575                           & 25                             & 9                            & 144                     \\
\textbf{BasicMotions}              & 80                            & 4                              & 6                            & 100                     \\
\textbf{CharacterTrajectories}     & 2858                          & 20                             & 3                            & 60-182                  \\
\textbf{Cricket}                   & 180                           & 12                             & 6                            & 1197                    \\
\textbf{Epilepsy}                  & 275                           & 4                              & 3                            & 206                     \\
\textbf{ERing}                     & 300                           & 6                              & 4                            & 65                      \\
\textbf{EthanolConcentration}      & 524                           & 4                              & 3                            & 1751                    \\
\textbf{EyesOpenShut}              & 98                            & 2                              & 14                           & 128                     \\
\textbf{FingerMovements}           & 416                           & 2                              & 28                           & 50                      \\
\textbf{Handwriting}               & 1000                          & 26                             & 3                            & 152                     \\
\textbf{JapaneseVowels}            & 640                           & 9                              & 12                           & 7-29                    \\
\textbf{Libras}                    & 360                           & 15                             & 2                            & 45                      \\
\textbf{NATOPS}                    & 360                           & 6                              & 24                           & 51                      \\
\textbf{RacketSports}              & 303                           & 4                              & 6                            & 30                      \\
\textbf{SpokenArabicDigits}        & 8798                          & 10                             & 13                           & 4-93                    \\ \bottomrule
\end{tabular}
\vspace{-1em}
\end{table}


\subsection{Details of datasets.}
We evaluated the proposed framework on a diverse collection of time series datasets from the University of East Anglia (UEA) and the University of California Riverside (UCR) Time Series Classification Repository\footnote{\url{http://www.timeseriesclassification.com/}}~\citep{bagnall_uea_2018,h_a_dau_ucr_2019}, utilizing the \textit{sktime} Python library \citep{loning_sktime_2019}. These repositories encompass a wide range of real-world applications, including both univariate and multivariate time series datasets with varying characteristics such as sample size, dimensionality, length, and the number of classes. 
As summarized in Table~\ref{tab:data}, we consider 30 different public datasets (15 univariate and 15 multivariate), each divided into three subsets: training, validation, and testing, following a 0.70/0.15/0.15 ratio. To address challenges of varying length time series within the datasets, we employed a uniform scaling approach as recommended by \citet{oh_stable_2024}. 

\paragraph{Data Preprocessing.} 
Our experiments were conducted on 30 diverse benchmark datasets, following the benchmark protocol established by  \citet{oh_stable_2024}. For each dataset, we introduced missing observations by randomly removing values, creating a more realistic scenario that mimics real-world data collection challenges, and considered four different missing rates, including regular (0\% missing), 30\%, 50\%, and 70\%. Furthermore, to investigate the robustness of the proposed framework, we performed five runs of cross-validation using different random seeds for each run.

\paragraph{NDE Architecture Design.} As backbone models for the proposed method, we utilize Neural ODE, Neural CDE, and Neural SDE, as explained in Section~\ref{sec:Methodology}. The vector field of the NDE models consists of nonlinear fully-connected layers with \texttt{ReLU} activation functions. 
Following the recommendation by \citet{kidger_neural_2020}, we included \texttt{Tanh} nonlinearity as the final operation for drift, diffusion, and all other vector fields to ensure stability.
For all methods, the (explicit) Euler method is employed as the ODE/SDE Solver. 

\paragraph{Hyperparameter optimization.}
To automate hyperparameter tuning  and minimize validation loss, we utilized the Python library, \textit{ray}\footnote{\url{https://github.com/ray-project/ray}}~\citep{moritz_ray_2018,liaw_tune_2018}. We first identified the best-performing hyperparameters on regular time series and then applied them to the corresponding irregular time series datasets.
We employed a combination of search strategies to explore the hyperparameter space. The learning rate $lr$ was optimized using a log-uniform search in the range of $10^{-4}$ to $10^{-1}$. For the number of layers $n_l$ in the vector field, we performed a grid search over the set $\left\{1,2,3,4\right\}$. Similarly, the dimensions of the hidden vectors $n_h$ were selected from the set $\left\{16,32,64,128\right\}$ using a grid search. The batch size was chosen from the set $\left\{16,32,64,128\right\}$, striking a balance between memory efficiency and convergence speed.

\begin{table*}[!htb]
\scriptsize\centering\captionsetup{skip=5pt}
\caption{Comparative analysis of performance of proposed method and baseline methods on the `GunPoint' dataset. Standard deviations of 5 runs are shown. Bold indicates best performance, underline indicates second-best. 
}\label{tab:case_result}
\begin{tabular}{@{}llccccccccc@{}}
\cmidrule(r){1-7} \cmidrule(l){9-11}
\multicolumn{1}{c}{\multirow{2.5}{*}{\textbf{Category}}} & \multicolumn{1}{c}{\multirow{2.5}{*}{\textbf{Method}}} & \multirow{2.5}{*}{\textbf{Regular}} & \multicolumn{3}{c}{\textbf{Missing Rate}}                                            & \multirow{2.5}{*}{\textbf{All settings}} & \multicolumn{1}{l}{} & \multicolumn{3}{c}{\textbf{Proposed method (with $\texttt{TANDEM}$}}                                                                       \\ \cmidrule(lr){4-6} \cmidrule(l){9-11} 
\multicolumn{1}{c}{}                                   & \multicolumn{1}{c}{}                                 &                                   & \textbf{30\%}               & \textbf{50\%}               & \textbf{70\%}               &                                        & \multicolumn{1}{l}{} & \multicolumn{1}{l}{\textbf{Neural ODE}} & \multicolumn{1}{l}{\textbf{Neural CDE}} & \multicolumn{1}{l}{\textbf{Neural SDE}} \\ \cmidrule(r){1-7} \cmidrule(l){9-11} 
\multirow{3}{*}{Imputation-based}                      & RNN                                                  & 0.547\std{0.132}                 & 0.480\std{0.069}          & 0.527\std{0.166}          & 0.553\std{0.051}          & 0.527\std{0.104}                      &                      & {20/0/0}(*)                             & {20/0/0}(*)                             & {20/0/0}(*)                             \\
                                                       & LSTM                                                 & 0.767\std{0.156}                 & 0.547\std{0.099}          & 0.513\std{0.051}          & 0.467\std{0.058}          & 0.573\std{0.091}                      &                      & {20/0/0}(*)                             & {20/0/0}(*)                             & {20/0/0}(*)                             \\
                                                       & GRU                                                  & 0.707\std{0.196}                 & 0.547\std{0.159}          & 0.547\std{0.080}          & 0.460\std{0.043}          & 0.565\std{0.120}                      &                      & {20/0/0}(*)                             & {20/0/0}(*)                             & {20/0/0}(*)                             \\ \cmidrule(r){1-7} \cmidrule(l){9-11} 
\multirow{3}{*}{Temporal feature}         & GRU-$\Delta t$                                       & 0.613\std{0.177}                 & 0.607\std{0.119}          & 0.593\std{0.064}          & 0.593\std{0.076}          & 0.602\std{0.109}                      &                      & {20/0/0}(*)                             & {20/0/0}(*)                             & {20/0/0}(*)                             \\
                                                       & GRU-Simple                                           & 0.693\std{0.095}                 & 0.567\std{0.175}          & 0.487\std{0.065}          & 0.493\std{0.028}          & 0.560\std{0.091}                      &                      & {20/0/0}(*)                             & {20/0/0}(*)                             & {20/0/0}(*)                             \\
                                                       & GRU-D                                                & 0.560\std{0.123}                 & 0.613\std{0.112}          & 0.560\std{0.083}          & 0.573\std{0.072}          & 0.577\std{0.098}                      &                      & {20/0/0}(*)                             & {20/0/0}(*)                             & {20/0/0}(*)                             \\ \cmidrule(r){1-7} \cmidrule(l){9-11} 
\multirow{3}{*}{Attention-based}                       & Transformer                                          & 0.487\std{0.045}                 & 0.487\std{0.115}          & 0.527\std{0.104}          & 0.453\std{0.110}          & 0.488\std{0.093}                      &                      & {20/0/0}(*)                             & {20/0/0}(*)                             & {20/0/0}(*)                             \\
                                                       & MTAN                                                 & 0.687\std{0.102}                 & 0.500\std{0.209}          & 0.573\std{0.119}          & 0.547\std{0.096}          & 0.577\std{0.131}                      &                      & {20/0/0}(*)                             & {20/0/0}(*)                             & {20/0/0}(*)                             \\
                                                       & MIAM                                                 & 0.540\std{0.055}                 & 0.527\std{0.080}          & 0.507\std{0.121}          & 0.533\std{0.058}          & 0.527\std{0.078}                      &                      & {20/0/0}(*)                             & {20/0/0}(*)                             & {20/0/0}(*)                             \\ \cmidrule(r){1-7} \cmidrule(l){9-11} 
\multirow{4}{*}{Neural   ODE-based}                    & Neural ODE                                           & 0.700\std{0.094}                 & 0.693\std{0.089}          & 0.693\std{0.076}          & 0.707\std{0.055}          & 0.698\std{0.079}                      &                      & {20/0/0}(*)                             & {20/0/0}(*)                             & {20/0/0}(*)                             \\
                                                       & GRU-ODE                                              & 0.913\std{0.030}                 & 0.920\std{0.051}          & 0.913\std{0.061}          & 0.920\std{0.069}          & 0.917\std{0.053}                      &                      & {12/5/3}(*)                             & {15/5/0}(*)                             & {14/3/3}(*)                             \\
                                                       & ODE-RNN                                              & 0.600\std{0.033}                 & 0.587\std{0.069}          & 0.607\std{0.136}          & 0.607\std{0.119}          & 0.600\std{0.089}                      &                      & {20/0/0}(*)                             & {20/0/0}(*)                             & {20/0/0}(*)                             \\
                                                       & ODE-LSTM                                             & 0.633\std{0.108}                 & 0.487\std{0.018}          & 0.500\std{0.085}          & 0.540\std{0.080}          & 0.540\std{0.073}                      &                      & {20/0/0}(*)                             & {20/0/0}(*)                             & {20/0/0}(*)                             \\ \cmidrule(r){1-7} \cmidrule(l){9-11} 
\multirow{5}{*}{Neural CDE-based}                      & Neural CDE                                           & \underline{0.973\std{0.028}}           & 0.940\std{0.028}          & 0.947\std{0.030}          & 0.927\std{0.043}          & 0.947\std{0.032}                      &                      & {7/9/4}                                 & {13/7/0}(*)                             & {12/4/4}                                \\
                                                       & Neural RDE                                           & 0.500\std{0.062}                 & 0.493\std{0.089}          & 0.493\std{0.076}          & 0.493\std{0.089}          & 0.495\std{0.079}                      &                      & {20/0/0}(*)                             & {20/0/0}(*)                             & {20/0/0}(*)                             \\
                                                       & ANCDE                                                & 0.587\std{0.141}                 & 0.693\std{0.109}          & 0.680\std{0.135}          & 0.653\std{0.161}          & 0.653\std{0.136}                      &                      & {19/1/0}(*)                             & {19/1/0}(*)                             & {19/0/1}(*)                             \\
                                                       & EXIT                                                 & 0.540\std{0.196}                 & 0.587\std{0.099}          & 0.507\std{0.132}          & 0.560\std{0.148}          & 0.548\std{0.144}                      &                      & {20/0/0}(*)                             & {20/0/0}(*)                             & {20/0/0}(*)                             \\
                                                       & LEAP                                                 & 0.460\std{0.043}                 & 0.533\std{0.085}          & 0.533\std{0.143}          & 0.580\std{0.084}          & 0.527\std{0.089}                      &                      & {20/0/0}(*)                             & {20/0/0}(*)                             & {20/0/0}(*)                             \\ \cmidrule(r){1-7} \cmidrule(l){9-11} 
\multirow{4}{*}{Neural   SDE-based}                    & Neural SDE                                           & 0.613\std{0.130}                 & 0.647\std{0.102}          & 0.673\std{0.086}          & 0.647\std{0.150}          & 0.645\std{0.117}                      &                      & {20/0/0}(*)                             & {20/0/0}(*)                             & {20/0/0}(*)                             \\
                                                       & Neural   LSDE                                        & 0.893\std{0.095}                 & 0.927\std{0.060}          & 0.893\std{0.060}          & 0.873\std{0.068}          & 0.897\std{0.071}                      &                      & {14/5/1}(*)                             & {19/0/1}(*)                             & {14/4/2}(*)                             \\
                                                       & Neural   LNSDE                                       & 0.887\std{0.084}                 & 0.880\std{0.077}          & 0.887\std{0.061}          & 0.900\std{0.075}          & 0.888\std{0.074}                      &                      & {14/5/1}(*)                             & {17/3/0}(*)                             & {16/1/3}(*)                             \\
                                                       & Neural   GSDE                                        & 0.853\std{0.096}                 & 0.867\std{0.062}          & 0.873\std{0.072}          & 0.873\std{0.068}          & 0.867\std{0.075}                      &                      & {17/2/1}(*)                             & {19/1/0}(*)                             & {18/1/1}(*)                             \\ \cmidrule(r){1-7} \cmidrule(l){9-11} 
\multirow{3}{*}{\textbf{Proposed}}                     & \textbf{Neural ODE (with $\texttt{TANDEM}$)}         & \underline{0.973\std{0.028}}           & 0.947\std{0.018}          & 0.967\std{0.033}          & \underline{0.960\std{0.037}}    & 0.962\std{0.029}                      &                      & -                                       & {10/8/2}(*)                             & {8/3/9}                                 \\
                                                       & \textbf{Neural CDE (with $\texttt{TANDEM}$)}       & \textbf{0.987\std{0.018}}        & \textbf{0.987\std{0.018}} & \textbf{0.980\std{0.030}} & \textbf{0.967\std{0.024}} & \textbf{0.980\std{0.022}}             &                      & {2/8/10}                                & -                                       & {4/6/10}                                \\
                                                       & \textbf{Neural SDE (with $\texttt{TANDEM}$)}       & \underline{0.973\std{0.015}}           & \underline{0.980\std{0.045}}    & \underline{0.973\std{0.028}}    & 0.933\std{0.033}          & \underline{0.965\std{0.030}}                &                      & {9/3/8}                                 & {10/6/4}(*)                             & -                                       \\ \cmidrule(r){1-7} \cmidrule(l){9-11} 
\end{tabular}
\vspace{-1em}
\end{table*}

\subsection{Comparative Baselines}
We evaluated our proposed method against a diverse set of baseline approaches, using their original source code and applying the same hyperparameter tuning procedure described earlier. The baselines can be categorized as follows:
\begin{itemize}
    \item \textbf{Imputation-based methods:} We employed mean imputation for recurrent neural network (RNN) \citep{rumelhart_learning_1986,medsker_recurrent_1999}, long short-term memory (LSTM) \citep{s_hochreiter_long_1997}, and gated recurrent unit (GRU) \citep{chung_empirical_2014} models to handle missingness.

    \item \textbf{Temporal feature incorporation:} Following \citet{choi_doctor_2016} and \citet{che_recurrent_2018}, we included models that incorporate additional temporal information, such as GRU-$\Delta t$ and GRU-Simple, which use time gaps and masks as additional inputs. Furthermore, GRU-D utilizes an exponential decay. 
    
    \item \textbf{Attention-based approaches:} We implemented the naïve Transformer \citep{vaswani_attention_2017} and enhanced attention-based frameworks such as multi-time attention networks (MTAN) \citep{shukla_multi-time_2021} and multi-integration attention module (MIAM) \citep{y_lee_multi-view_2022} to leverage attention mechanisms. 
        
    \item \textbf{Neural ODE-based methods:} We compared our approach with various neural ODE-based implementations, including GRU-ODE \citep{brouwer_gru-ode-bayes_2019}, ODE-RNN \citep{rubanova_latent_2019}, and ODE-LSTM \citep{lechner_learning_2020}. 
    
    \item \textbf{Neural CDE-based approaches:} We used the original Neural CDE with Hermite cubic splines \citep{kidger_neural_2020}, Neural RDE with depth 2 signature transform and cubic interpolation. 
    Additional to that, ANCDE~\citep{jhin_attentive_2024}, EXIT (EXtrapolation and InTerpolation)~\citep{jhin_exit_2022}, and LEAP (LEArnable Path)~\citep{jhin_learnable_2023} are included.

    \item \textbf{Neural SDE-based methods:} We considered statistically stable Neural SDEs proposed by \citet{oh_stable_2024}: Neural Langevin-type SDE (LSDE), Neural Linear Noise SDE (LNSDE), and Neural Geometric SDE (GSDE). These models combine control path into the augmented states.  
\end{itemize}

\subsection{Case Study: `GunPoint' Dataset}
The `GunPoint' dataset, introduced by \citet{ratanamahatana_three_2005}, is a binary classification problem in time series analysis. It consists of motion data from two actors (one female, one male) performing two distinct hand movements: Gun-Draw and Point. The dataset tracks the X-axis movement of the actors' right hands, capturing the temporal dynamics of these actions. 
To further illustrate the effectiveness of our proposed \texttt{TANDEM} framework, we conducted a detailed case study, under four different settings.

Our \texttt{TANDEM} framework demonstrates robust performance across all settings, maintaining high accuracy even with significant portions of missing data, as represented in Table~\ref{tab:case_result}. 
When missing rates are increased, the benchmark's performance degrades, while our model remains impressively high accuracy. 
The integration of $\texttt{TANDEM}$ yields performance improvements across all baselines. We observe significant enhancements in Neural ODE and Neural SDE, whereas the improvements in Neural CDE are more moderate.

To rigorously assess performance differences between \texttt{TANDEM} and baseline methods, we employ the one-sided Wilcoxon signed-rank test \citep{demsar_statistical_2006}, suitable for comparing paired results on the same dataset. To account for multiple comparisons, we apply the Holm-Bonferroni correction \citep{giacalone_bonferroni-holm_2018}, using a corrected significance threshold of $p<0.05$ in line with best practices \citep{benavoli_should_2016}.
We analyze `GunPoint' dataset individually across four missing rates (0\%, 30\%, 50\%, 70\%), with five repetitions per condition, yielding 20 paired observations per model. In the right panel of Table~\ref{tab:case_result}, our framework clearly outperforms the baseline methods, with Neural CDE being the strongest competitor. However, our approach (Neural CDE with \texttt{TANDEM}) achieves a slight improvement even over this strong baseline.
Although this single-dataset analysis provides useful insights, its generalizability is limited. To address this, we extend our experiments to a broader set of datasets as represented in Table~\ref{tab:data}.

\begin{table*}[!htb]
\scriptsize\centering\captionsetup{skip=5pt}
\caption{Classification performance on 30 benchmark datasets with regular and three missing rates. The values within the parentheses indicate the average of individual standard deviations. 
(*) denotes statistically significant superiority of pairwise comparison (Win/Tie/Loss counts) and one-sided Wilcoxon signed-rank test with $p<0.05$
}\label{tab:result_full}
\begin{tabular}{@{}lC{0.8cm}C{0.5cm}C{0.8cm}C{0.5cm}C{0.8cm}C{0.5cm}C{0.8cm}C{0.5cm}C{0.8cm}C{0.5cm}cccc@{}}
\cmidrule(r){1-11} \cmidrule(l){13-15}
\multicolumn{1}{c}{\multirow{2.5}{*}{\textbf{Method}}} & \multicolumn{2}{c}{\textbf{Regular}}         & \multicolumn{2}{c}{\textbf{30\% Missing}}    & \multicolumn{2}{c}{\textbf{50\% Missing}}    & \multicolumn{2}{c}{\textbf{70\% Missing}}    & \multicolumn{2}{c}{\textbf{All settings}} &  & \multicolumn{3}{c}{\textbf{Proposed method (with $\texttt{TANDEM}$)}}           \\ \cmidrule(lr){2-11} \cmidrule(l){13-15} 
\multicolumn{1}{c}{}                                 & \textbf{Accuracy}            & \textbf{Rank} & \textbf{Accuracy}            & \textbf{Rank} & \textbf{Accuracy}            & \textbf{Rank} & \textbf{Accuracy}            & \textbf{Rank} & \textbf{Accuracy}      & \textbf{Rank}    &  & \textbf{Neural ODE} & \textbf{Neural CDE} & \textbf{Neural SDE} \\ \cmidrule(r){1-11} \cmidrule(l){13-15} 
RNN                                                  & 0.582\avgstd{0.064}          & 18.42         & 0.513\avgstd{0.087}          & 20.12         & 0.485\avgstd{0.088}          & 21.70         & 0.472\avgstd{0.072}          & 20.58         & 0.513\avgstd{0.078}    & 20.20            &  & {114/1/5}(*)        & {113/2/5}(*)        & {114/0/6}(*)        \\
LSTM                                                 & 0.633\avgstd{0.053}          & 15.40         & 0.595\avgstd{0.060}          & 16.33         & 0.567\avgstd{0.061}          & 17.15         & 0.558\avgstd{0.058}          & 16.60         & 0.589\avgstd{0.058}    & 16.37            &  & {99/6/15}(*)        & {104/0/16}(*)       & {108/0/12}(*)       \\
GRU                                                  & 0.672\avgstd{0.059}          & 11.45         & 0.621\avgstd{0.063}          & 13.77         & 0.610\avgstd{0.055}          & 13.97         & 0.597\avgstd{0.062}          & 13.78         & 0.625\avgstd{0.060}    & 13.24            &  & {81/3/36}(*)        & {90/2/28}(*)        & {85/4/31}(*)        \\ \cmidrule(r){1-11} \cmidrule(l){13-15} 
GRU-$\Delta t$                                       & 0.641\avgstd{0.070}          & 14.00         & 0.636\avgstd{0.066}          & 12.37         & 0.634\avgstd{0.056}          & 11.93         & 0.618\avgstd{0.065}          & 13.75         & 0.632\avgstd{0.064}    & 13.01            &  & {81/4/35}(*)        & {85/2/33}(*)        & {84/5/31}(*)        \\
GRU-Simple                                           & 0.669\avgstd{0.060}          & 11.38         & 0.640\avgstd{0.063}          & 11.55         & 0.613\avgstd{0.071}          & 13.62         & 0.569\avgstd{0.072}          & 15.78         & 0.623\avgstd{0.066}    & 13.08            &  & {83/3/34}(*)        & {86/1/33}(*)        & {85/1/34}(*)        \\
GRU-D                                                & 0.648\avgstd{0.071}          & 14.32         & 0.624\avgstd{0.075}          & 14.35         & 0.611\avgstd{0.073}          & 15.07         & 0.604\avgstd{0.067}          & 14.38         & 0.622\avgstd{0.072}    & 14.53            &  & {93/0/27}(*)        & {98/4/18}(*)        & {92/3/25}(*)        \\ \cmidrule(r){1-11} \cmidrule(l){13-15} 
Transformer                                          & 0.664\avgstd{0.069}          & 13.52         & 0.625\avgstd{0.068}          & 15.05         & 0.627\avgstd{0.061}          & 14.97         & 0.607\avgstd{0.066}          & 15.02         & 0.631\avgstd{0.066}    & 14.64            &  & {90/4/26}(*)        & {92/1/27}(*)        & {96/2/22}(*)        \\
MTAN                                                 & 0.648\avgstd{0.080}          & 16.42         & 0.618\avgstd{0.099}          & 14.80         & 0.618\avgstd{0.091}          & 13.82         & 0.607\avgstd{0.078}          & 13.30         & 0.623\avgstd{0.087}    & 14.58            &  & {97/1/22}(*)        & {102/1/17}(*)       & {96/1/23}(*)        \\
MIAM                                                 & 0.623\avgstd{0.048}          & 14.88         & 0.603\avgstd{0.066}          & 14.92         & 0.589\avgstd{0.063}          & 16.58         & 0.569\avgstd{0.056}          & 16.38         & 0.596\avgstd{0.058}    & 15.69            &  & {95/3/22}(*)        & {99/2/19}(*)        & {93/4/23}(*)        \\ \cmidrule(r){1-11} \cmidrule(l){13-15} 
Neural ODE                                           & 0.521\avgstd{0.065}          & 18.70         & 0.518\avgstd{0.061}          & 17.47         & 0.515\avgstd{0.060}          & 17.22         & 0.526\avgstd{0.058}          & 16.28         & 0.520\avgstd{0.061}    & 17.42            &  & {112/3/5}(*)        & {109/0/11}(*)       & {108/5/7}(*)        \\
GRU-ODE                                              & 0.671\avgstd{0.067}          & 13.72         & 0.663\avgstd{0.064}          & 13.23         & 0.666\avgstd{0.059}          & 11.52         & 0.655\avgstd{0.062}          & 11.22         & 0.664\avgstd{0.063}    & 12.42            &  & {93/1/26}(*)        & {96/2/22}(*)        & {86/1/33}(*)        \\
ODE-RNN                                              & 0.658\avgstd{0.063}          & 12.73         & 0.635\avgstd{0.064}          & 13.13         & 0.636\avgstd{0.067}          & 11.38         & 0.630\avgstd{0.055}          & 11.55         & 0.640\avgstd{0.062}    & 12.20            &  & {87/3/30}(*)        & {90/3/27}(*)        & {83/5/32}(*)        \\
ODE-LSTM                                             & 0.619\avgstd{0.063}          & 15.42         & 0.584\avgstd{0.064}          & 16.30         & 0.561\avgstd{0.065}          & 17.60         & 0.530\avgstd{0.085}          & 17.20         & 0.574\avgstd{0.069}    & 16.63            &  & {99/3/18}(*)        & {105/0/15}(*)       & {102/3/15}(*)       \\ \cmidrule(r){1-11} \cmidrule(l){13-15} 
Neural CDE                                           & 0.709\avgstd{0.061}          & 12.03         & 0.706\avgstd{0.073}          & 9.47          & 0.696\avgstd{0.064}          & 9.42          & 0.665\avgstd{0.072}          & 10.93         & 0.694\avgstd{0.068}    & 10.46            &  & {75/5/40}(*)        & {88/6/26}(*)        & {76/3/41}(*)        \\
Neural RDE                                           & 0.678\avgstd{0.066}          & 13.02         & 0.658\avgstd{0.074}          & 12.57         & 0.640\avgstd{0.067}          & 13.65         & 0.626\avgstd{0.077}          & 12.33         & 0.651\avgstd{0.071}    & 12.89            &  & {83/4/33}(*)        & {95/4/21}(*)        & {85/1/34}(*)        \\
ANCDE                                                & 0.693\avgstd{0.067}          & 11.23         & 0.687\avgstd{0.068}          & 10.60         & 0.683\avgstd{0.078}          & 9.82          & 0.655\avgstd{0.067}          & 10.07         & 0.679\avgstd{0.070}    & 10.43            &  & {74/2/44}(*)        & {89/4/27}(*)        & {77/3/40}(*)        \\
EXIT                                                 & 0.636\avgstd{0.073}          & 15.37         & 0.633\avgstd{0.078}          & 14.60         & 0.616\avgstd{0.075}          & 14.68         & 0.599\avgstd{0.075}          & 15.28         & 0.621\avgstd{0.075}    & 14.98            &  & {103/2/15}(*)       & {106/1/13}(*)       & {105/1/14}(*)       \\
LEAP                                                 & 0.444\avgstd{0.068}          & 19.60         & 0.401\avgstd{0.078}          & 21.07         & 0.425\avgstd{0.073}          & 19.35         & 0.414\avgstd{0.070}          & 19.73         & 0.421\avgstd{0.072}    & 19.94            &  & {115/0/5}(*)        & {113/0/7}(*)        & {111/1/8}(*)        \\ \cmidrule(r){1-11} \cmidrule(l){13-15} 
Neural   SDE                                         & 0.526\avgstd{0.068}          & 17.77         & 0.508\avgstd{0.066}          & 17.67         & 0.517\avgstd{0.058}          & 17.68         & 0.512\avgstd{0.066}          & 17.68         & 0.516\avgstd{0.064}    & 17.70            &  & {109/0/11}(*)       & {106/1/13}(*)       & {107/1/12}(*)       \\
Neural   LSDE                                        & 0.717\avgstd{0.056}          & 7.93          & 0.690\avgstd{0.050}          & 9.23          & 0.686\avgstd{0.051}          & 8.72          & 0.682\avgstd{0.067}          & 7.30          & 0.694\avgstd{0.056}    & 8.30             &  & {62/8/50}(*)        & {72/3/45}(*)        & {61/11/48}(*)       \\
Neural   LNSDE                                       & 0.727\avgstd{0.047}          & 7.98          & 0.723\avgstd{0.050}          & 7.48          & 0.717\avgstd{0.054}          & \textbf{6.20} & \underline{0.703\avgstd{0.054}}    & \textbf{6.13} & 0.717\avgstd{0.051}    & \underline{6.95}       &  & {59/6/55}           & {60/4/56}           & {52/6/62}           \\
Neural   GSDE                                        & 0.716\avgstd{0.065}          & 8.22          & 0.707\avgstd{0.069}          & 8.02          & 0.698\avgstd{0.063}          & 8.43          & 0.689\avgstd{0.056}          & 7.67          & 0.703\avgstd{0.063}    & 8.08             &  & {59/8/53}           & {71/4/45}(*)        & {59/7/54}           \\ \cmidrule(r){1-11} \cmidrule(l){13-15} 
\textbf{Neural ODE (with   $\texttt{TANDEM}$)}       & 0.738\avgstd{0.055}          & 8.70          & \underline{0.737\avgstd{0.063}}    & \underline{6.67}    & \underline{0.737\avgstd{0.060}}    & 6.52          & 0.700\avgstd{0.067}          & 7.62          & 0.728\avgstd{0.061}    & 7.38             &  & -                   & {65/3/52}           & {51/11/58}          \\
\textbf{Neural CDE (with   $\texttt{TANDEM}$)}       & \textbf{0.767\avgstd{0.053}} & \textbf{5.87} & \textbf{0.750\avgstd{0.051}} & \textbf{6.58} & \textbf{0.742\avgstd{0.062}} & \underline{6.37}    & \textbf{0.709\avgstd{0.064}} & \underline{6.83}    & 0.742\avgstd{0.057}    & \textbf{6.41}    &  & {52/3/65}           & -                   & {48/2/70}           \\
\textbf{Neural SDE (with   $\texttt{TANDEM}$)}       & \underline{0.749\avgstd{0.058}}    & \underline{6.93}    & 0.724\avgstd{0.054}          & 7.67          & 0.713\avgstd{0.060}          & 7.65          & 0.696\avgstd{0.051}          & 7.58          & 0.721\avgstd{0.056}    & 7.46             &  & {58/11/51}          & {70/2/48}(*)        & -                   \\ \cmidrule(r){1-11} \cmidrule(l){13-15} 
\end{tabular}
\vspace{-1em}
\end{table*}

\begin{figure}[!htb]
    \centering\captionsetup{skip=5pt}
    \captionsetup[subfigure]{justification=centering, skip=5pt}
    \subfloat[Accuracy score versus $\log_{10}$ number of parameters]{
      \includegraphics[width=0.48\linewidth]{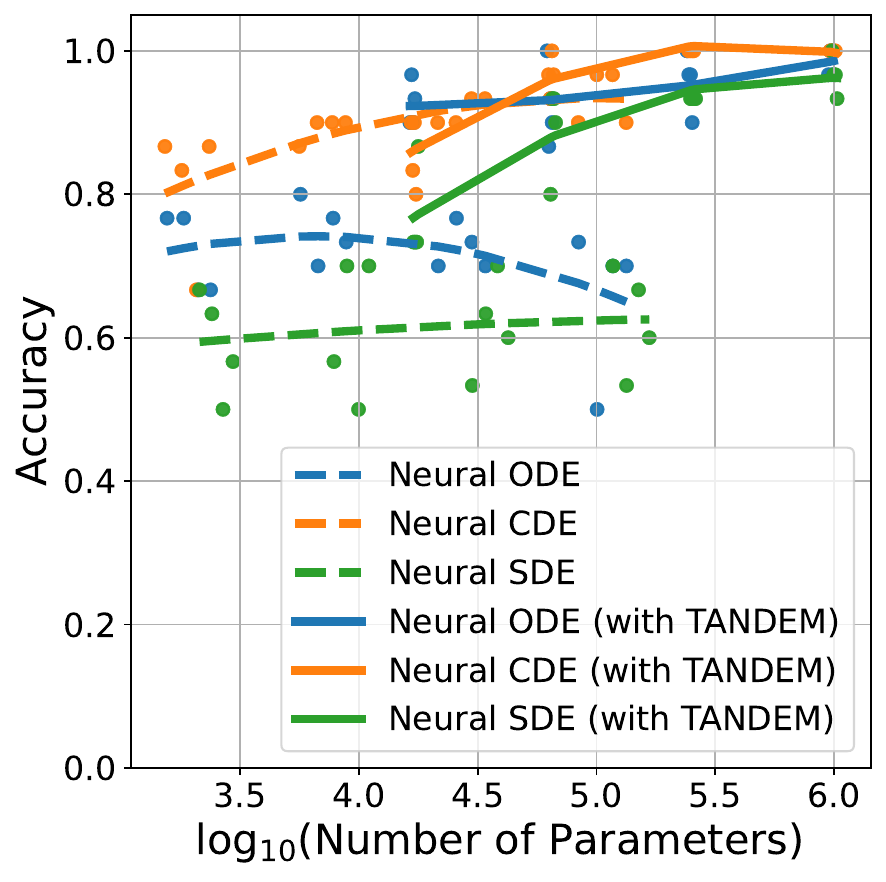}} \hfil
    \subfloat[Accuracy score versus runtime per epoch in seconds]{
      \includegraphics[width=0.48\linewidth]{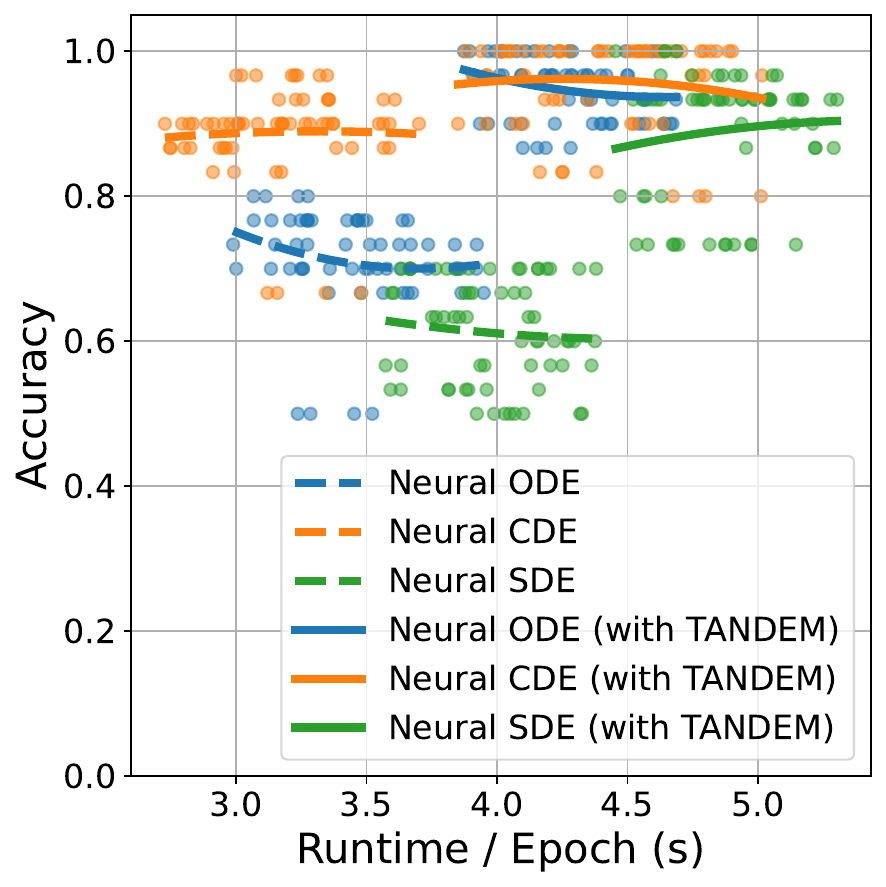}}
    \caption{Performance-computation trade-off analysis with NDE baselines and their \texttt{TANDEM} variants. Each point reflects a model with different layer counts ($n_l$) and hidden sizes ($n_h$). Dashed/Solid lines show second-order polynomial trends for each model family. Each configuration is repeated five times.
    }
    \label{fig:computation}
    \Description{}
\end{figure}

For further analysis, we examine the performance–computation trade-off of our proposed method, as shown in Figure~\ref{fig:computation}.
Note that computational time is largely influenced by the choice of solver~\citep{oh_stable_2024}. 
Specifically, we used \textit{torchcde}\footnote{\url{https://github.com/patrick-kidger/torchcde}}~\citep{kidger_neural_2020} for Neural CDEs, and \textit{torchsde}\footnote{\url{https://github.com/google-research/torchsde}}~\citep{li_scalable_2020} for Neural ODEs and SDEs.

Figure~\ref{fig:computation}~(a) and (b) together illustrate the trade-off between accuracy and computation. While Figure~\ref{fig:computation}~(a) shows that \texttt{TANDEM} improves parameter efficiency by achieving higher accuracy with comparable model sizes, Figure~\ref{fig:computation}~(b) reveals the cost: \texttt{TANDEM} variants consistently require longer runtimes per epoch due to added attention and gating overhead. This highlights a key trade-off between predictive performance and computational efficiency.

\subsection{Comprehensive results on 30 Datasets}
Table~\ref{tab:result_full} compares the classification accuracy over five iterations. 
We observe that \texttt{TANDEM} maintains relatively stable performance across all tested scenarios, in contrast to most baselines that degrade noticeably under higher missingness. This stability suggests that the proposed incorporation of explicit features provides a robust mechanism for recovering meaningful temporal dynamics, even in situations where large fractions of the data are unobserved. 

For a broader comparison, we average results over five runs for each of the 30 datasets and four missing rates, generating 120 data points per method. These are used for pairwise tests and corrected $p$-values, offering a more robust evaluation across diverse datasets. The number of pairwise comparisons (including Win/Tie/Loss counts) and statistical test results are summarized in the right panel of Table~ \ref{tab:result_full}. 
For instance, Neural CDE with TANDEM achieves the best average rank across all missingness settings. The Win/Tie/Loss counts further highlight the TANDEM framework's effectiveness, showing numerous statistically significant wins against other approaches. 
When compared to Neural LNSDE, our method achieves more wins in one-to-one comparisons, although the difference is not statistically significant.

\begin{table}[htb]
\scriptsize\centering\captionsetup{skip=5pt}
\caption{Average value of the proposed gate activations on 30 benchmark datasets for varying missing rates (The values within the parentheses indicate the average of individual standard deviations.)}\label{tab:gating}
\begin{tabular}{@{}ccccc@{}}
\toprule
\multirow{2.5}{*}{\textbf{\begin{tabular}[c]{@{}c@{}}Neural   ODE \\      (with $\texttt{TANDEM}$)\end{tabular}}} & \multirow{2.5}{*}{\textbf{Regular}} & \multicolumn{3}{c}{\textbf{Missing Rate}}                       \\ \cmidrule(l){3-5} 
                                                                                                                &                                   & \textbf{30\%}       & \textbf{50\%}       & \textbf{70\%}       \\ \midrule
$\sigma_{\tilde{\vx}}$                                                                                          & 0.473\avgstd{0.393}               & 0.513\avgstd{0.424} & 0.447\avgstd{0.411} & 0.473\avgstd{0.357} \\
$\sigma_{\bm{X}}$                                                                                               & 0.533\avgstd{0.408}               & 0.520\avgstd{0.386} & 0.580\avgstd{0.351} & 0.527\avgstd{0.413} \\
$\sigma_{\vz}$                                                                                                  & 0.880\avgstd{0.111}               & 0.853\avgstd{0.158} & 0.873\avgstd{0.118} & 0.867\avgstd{0.192} \\ \bottomrule
\end{tabular}
\\[\baselineskip]
\begin{tabular}{@{}ccccc@{}}
\toprule
\multirow{2.5}{*}{\textbf{\begin{tabular}[c]{@{}c@{}}Neural   CDE \\      (with $\texttt{TANDEM}$)\end{tabular}}} & \multirow{2.5}{*}{\textbf{Regular}} & \multicolumn{3}{c}{\textbf{Missing Rate}}                       \\ \cmidrule(l){3-5} 
                                                                                                                &                                   & \textbf{30\%}       & \textbf{50\%}       & \textbf{70\%}       \\ \midrule
$\sigma_{\tilde{\vx}}$                                                                                          & 0.340\avgstd{0.353}               & 0.307\avgstd{0.365} & 0.273\avgstd{0.342} & 0.307\avgstd{0.443} \\
$\sigma_{\bm{X}}$                                                                                               & 0.187\avgstd{0.270}               & 0.233\avgstd{0.376} & 0.220\avgstd{0.277} & 0.280\avgstd{0.365} \\
$\sigma_{\vz}$                                                                                                  & 0.720\avgstd{0.396}               & 0.713\avgstd{0.345} & 0.680\avgstd{0.398} & 0.707\avgstd{0.353} \\ \bottomrule
\end{tabular}
\\[\baselineskip]
\begin{tabular}{@{}ccccc@{}}
\toprule
\multirow{2.5}{*}{\textbf{\begin{tabular}[c]{@{}c@{}}Neural   SDE \\      (with $\texttt{TANDEM}$)\end{tabular}}} & \multirow{2.5}{*}{\textbf{Regular}} & \multicolumn{3}{c}{\textbf{Missing Rate}}                       \\ \cmidrule(l){3-5} 
                                                                                                                &                                   & \textbf{30\%}       & \textbf{50\%}       & \textbf{70\%}       \\ \midrule
$\sigma_{\tilde{\vx}}$                                                                                          & 0.533\avgstd{0.340}               & 0.487\avgstd{0.372} & 0.500\avgstd{0.313} & 0.447\avgstd{0.395} \\
$\sigma_{\bm{X}}$                                                                                               & 0.487\avgstd{0.309}               & 0.527\avgstd{0.365} & 0.533\avgstd{0.317} & 0.527\avgstd{0.353} \\
$\sigma_{\vz}$                                                                                                  & 0.867\avgstd{0.144}               & 0.847\avgstd{0.169} & 0.833\avgstd{0.194} & 0.853\avgstd{0.166} \\ \bottomrule
\end{tabular}
\end{table}

Furthermore, the results show that our attention-guided framework significantly improves performance for both neural ODE and SDE models compared to their na\"ive approaches. Neural CDE shows modest performance gains with proposed method since it already incorporates control paths in its formulation. Nevertheless, Neural CDE with $\texttt{TANDEM}$ achieves the highest accuracy across all scenarios.
This improvement supports our hypothesis that incorporating complementary inputs through attention mechanisms can enhance the model's ability to handle missing data in time series.

Table~\ref{tab:gating} presents the average gating values $\sigma_{\tilde{\vx}}$, $\sigma_{\bm{X}}$, and $\sigma_{\vz}$ across 30 datasets and various missing rates. While each individual gate is drawn from a near-binary \texttt{Gumbel-Sigmoid} distribution at the level of a single sample and time step, the table reports averages over large collections of runs. 

A key takeaway is that when missingness is moderate to high, $\sigma_X$ and $\sigma_z$ often increase, suggesting that the gating mechanism elevates interpolation or latent trajectory in regions of sparse or noisy measurements. Conversely, at lower missing rates, $\sigma_{\tilde{x}}$ remains non-trivial, indicating that direct observations provide a concrete information whenever they are sufficiently present. 
In conclusion, these average gate values align well with the intuition that \texttt{TANDEM} adaptively allocates more weight to the control path or latent dynamics when raw data are unreliable, yet reverts to directly using raw measurements when they are abundant and trustworthy.

\subsection{Ablation Study}

\begin{table}[!htb]
\scriptsize\centering\captionsetup{skip=5pt}
\caption{Ablation study on 30 benchmark datasets with regular and three missing rates  (The values within the parentheses indicate the average of individual standard deviations.)
}\label{tab:ablation_full}
\begin{tabular}{@{}clC{0.8cm}C{0.8cm}C{0.8cm}C{0.8cm}C{0.8cm}@{}}
\toprule
\multirow{2.5}{*}{~} & \multicolumn{1}{c}{\multirow{2.5}{*}{\textbf{Component}}} & \multirow{2.5}{*}{\textbf{Regular}} & \multicolumn{3}{c}{\textbf{Missing Rate}}                       & \multirow{2.5}{*}{\textbf{\begin{tabular}[c]{@{}c@{}}All\\ Settings\end{tabular}}} \\ \cmidrule(lr){4-6}
                                     & \multicolumn{1}{c}{}                                    &                                   & \textbf{30\%}        & \textbf{50\%}        & \textbf{70\%}        &                                        \\ \midrule
\multirow{13}{*}{\rotatebox{90}{\textbf{Neural ODE}}} & \texttt{TANDEM}                                         & 0.738 \avgstd{0.055}               & 0.737 \avgstd{0.063} & 0.737 \avgstd{0.060} & 0.700 \avgstd{0.067} & 0.728 \avgstd{0.061}                    \\
                                     & \texttt{TANDEM} (soft)                                  & 0.706 \avgstd{0.063}               & 0.701 \avgstd{0.075} & 0.681 \avgstd{0.071} & 0.671 \avgstd{0.066} & 0.690 \avgstd{0.068}                    \\
                                     & $[\Phi_{z(t)}$,   $\Phi_{\tilde{x}(t)}$, $\Phi_{X(t)}]$ & 0.701 \avgstd{0.063}               & 0.687 \avgstd{0.067} & 0.670 \avgstd{0.073} & 0.652 \avgstd{0.072} & 0.677 \avgstd{0.069}                    \\
                                     & $[\Phi_{z(t)}$, $\Phi_{X(t)}]$                          & 0.710 \avgstd{0.063}               & 0.701 \avgstd{0.068} & 0.690 \avgstd{0.080} & 0.673 \avgstd{0.070} & 0.694 \avgstd{0.070}                    \\
                                     & $[\Phi_{z(t)}$,   $\Phi_{\tilde{x}(t)}]$                & 0.688 \avgstd{0.069}               & 0.681 \avgstd{0.075} & 0.658 \avgstd{0.072} & 0.636 \avgstd{0.078} & 0.666 \avgstd{0.073}                    \\
                                     & $\Phi_{z(t)}$ only                                      & 0.519 \avgstd{0.074}               & 0.520 \avgstd{0.060} & 0.516 \avgstd{0.064} & 0.525 \avgstd{0.062} & 0.520 \avgstd{0.065}                    \\
                                     & w/o attention                                           & 0.521 \avgstd{0.065}               & 0.518 \avgstd{0.061} & 0.515 \avgstd{0.060} & 0.526 \avgstd{0.058} & 0.520 \avgstd{0.061}                    \\ \midrule
\multirow{13}{*}{\rotatebox{90}{\textbf{Neural CDE}}} & \texttt{TANDEM}                                         & 0.767 \avgstd{0.053}               & 0.750 \avgstd{0.051} & 0.742 \avgstd{0.062} & 0.709 \avgstd{0.064} & 0.742 \avgstd{0.057}                    \\
                                     & \texttt{TANDEM} (soft)                                  & 0.751 \avgstd{0.057}               & 0.737 \avgstd{0.063} & 0.722 \avgstd{0.065} & 0.692 \avgstd{0.070} & 0.726 \avgstd{0.063}                    \\
                                     & $[\Phi_{z(t)}$,   $\Phi_{\tilde{x}(t)}$, $\Phi_{X(t)}]$ & 0.720 \avgstd{0.057}               & 0.714 \avgstd{0.071} & 0.708 \avgstd{0.066} & 0.684 \avgstd{0.068} & 0.707 \avgstd{0.066}                    \\
                                     & $[\Phi_{z(t)}$, $\Phi_{X(t)}]$                          & 0.711 \avgstd{0.059}               & 0.710 \avgstd{0.067} & 0.692 \avgstd{0.058} & 0.677 \avgstd{0.066} & 0.697 \avgstd{0.062}                    \\
                                     & $[\Phi_{z(t)}$,   $\Phi_{\tilde{x}(t)}]$                & 0.741 \avgstd{0.055}               & 0.727 \avgstd{0.060} & 0.703 \avgstd{0.075} & 0.681 \avgstd{0.064} & 0.713 \avgstd{0.064}                    \\
                                     & $\Phi_{z(t)}$ only                                      & 0.724 \avgstd{0.073}               & 0.707 \avgstd{0.055} & 0.702 \avgstd{0.067} & 0.684 \avgstd{0.063} & 0.705 \avgstd{0.065}                    \\
                                     & w/o attention                                           & 0.709 \avgstd{0.061}               & 0.706 \avgstd{0.073} & 0.696 \avgstd{0.064} & 0.665 \avgstd{0.072} & 0.694 \avgstd{0.068}                    \\ \midrule
\multirow{13}{*}{\rotatebox{90}{\textbf{Neural SDE}}} & \texttt{TANDEM}                                         & 0.749 \avgstd{0.058}               & 0.724 \avgstd{0.054} & 0.713 \avgstd{0.060} & 0.696 \avgstd{0.051} & 0.721 \avgstd{0.056}                    \\
                                     & \texttt{TANDEM} (soft)                                  & 0.718 \avgstd{0.067}               & 0.689 \avgstd{0.060} & 0.677 \avgstd{0.063} & 0.659 \avgstd{0.066} & 0.686 \avgstd{0.064}                    \\
                                     & $[\Phi_{z(t)}$,   $\Phi_{\tilde{x}(t)}$, $\Phi_{X(t)}]$ & 0.709 \avgstd{0.051}               & 0.689 \avgstd{0.063} & 0.669 \avgstd{0.072} & 0.657 \avgstd{0.061} & 0.681 \avgstd{0.062}                    \\
                                     & $[\Phi_{z(t)}$, $\Phi_{X(t)}]$                          & 0.677 \avgstd{0.078}               & 0.663 \avgstd{0.076} & 0.655 \avgstd{0.070} & 0.655 \avgstd{0.060} & 0.662 \avgstd{0.071}                    \\
                                     & $[\Phi_{z(t)}$,   $\Phi_{\tilde{x}(t)}]$                & 0.687 \avgstd{0.067}               & 0.656 \avgstd{0.071} & 0.645 \avgstd{0.075} & 0.637 \avgstd{0.071} & 0.656 \avgstd{0.071}                    \\
                                     & $\Phi_{z(t)}$ only                                      & 0.524 \avgstd{0.059}               & 0.523 \avgstd{0.068} & 0.520 \avgstd{0.063} & 0.524 \avgstd{0.068} & 0.523 \avgstd{0.064}                    \\
                                     & w/o attention                                           & 0.526 \avgstd{0.068}               & 0.508 \avgstd{0.066} & 0.517 \avgstd{0.058} & 0.512 \avgstd{0.066} & 0.516 \avgstd{0.064}                    \\ \bottomrule
\end{tabular}
\vspace{-1em}
\end{table}

We evaluated the performance of the framework using various combinations of attention components, including raw observation $\tilde{\vx}(t)$, control path $\bm{X}(t)$, and continuous latent dynamics $\vz(t)$. 
We also considered soft method without hard thresholding in the attention gating module. The na\"ive approach, which did not include any attention mechanism, served as a baseline for comparison in Table~\ref{tab:ablation_full}.
Overall, \texttt{TANDEM} framework improves the performance across various benchmark datasets with missing values. 

\begin{figure}[!htb]
    \centering\captionsetup{skip=5pt}
    \captionsetup[subfigure]{justification=centering, skip=5pt}
    \subfloat[Neural ODE]{
      \includegraphics[width=0.95\linewidth]{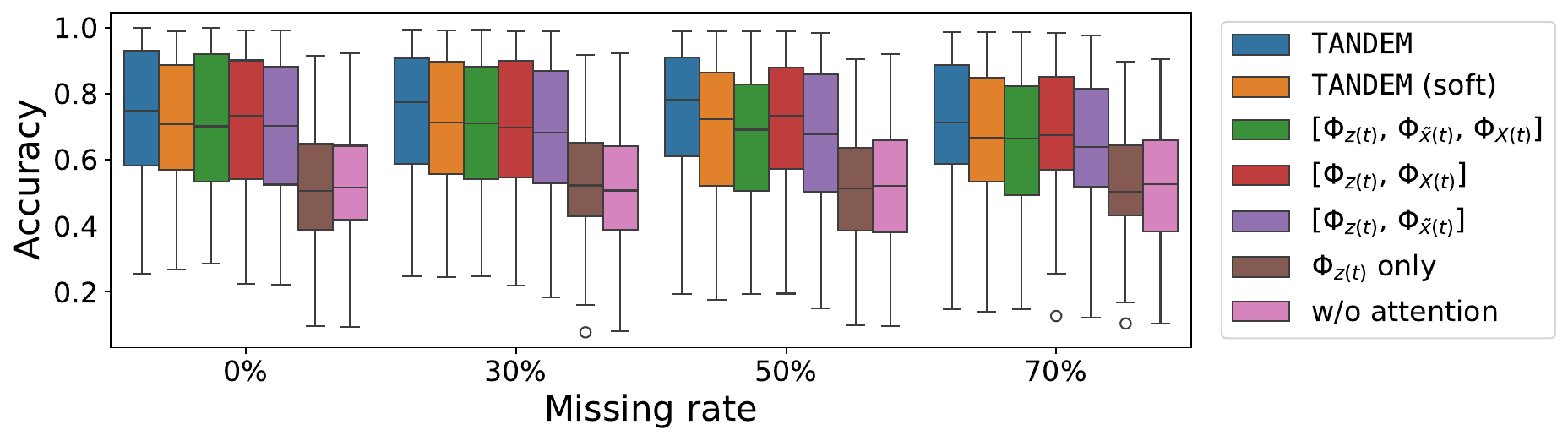}} \\ [0.5em]
    \subfloat[Neural CDE]{
      \includegraphics[width=0.95\linewidth]{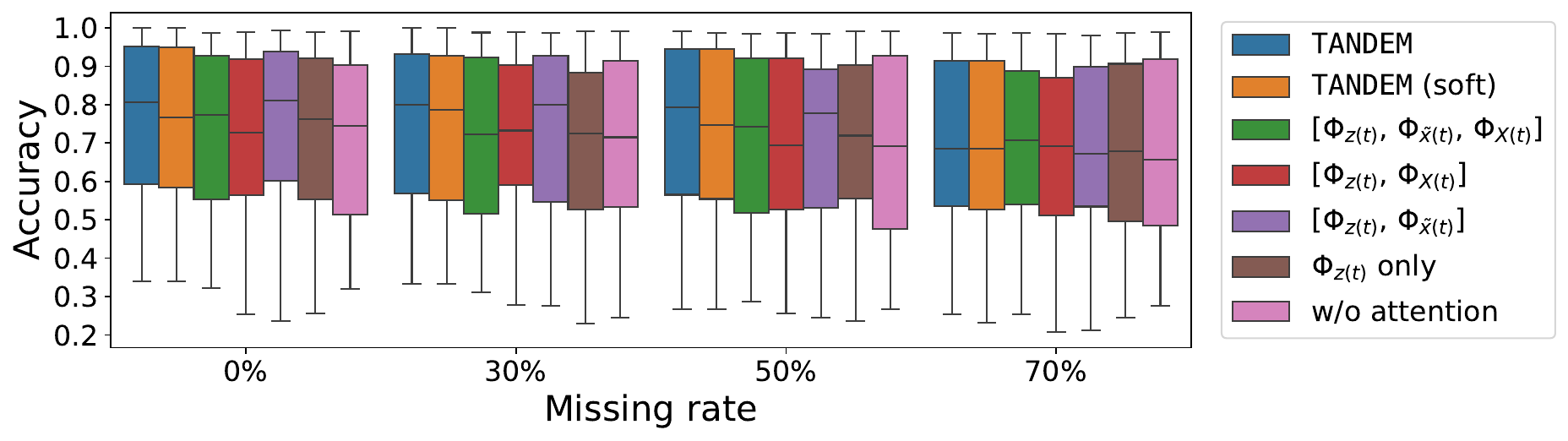}} \\ [0.5em]
    \subfloat[Neural SDE]{
      \includegraphics[width=0.95\linewidth]{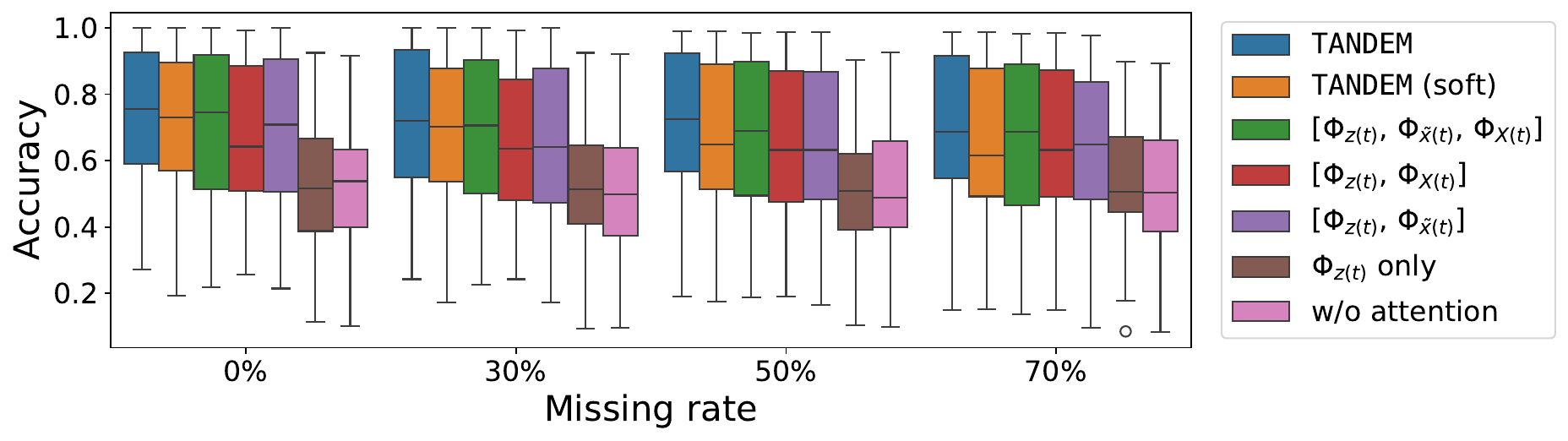}}
    \caption{Ablation study regarding model configuration 
    }
    \label{fig:ablation}
    \Description{}
\end{figure}

Figure~\ref{fig:ablation} then summarizes an ablation study, showing how different feature components affect performance. We observe that removing attention or restricting the model to a single feature yields a drop in accuracy across all missing rates, confirming the importance of multi-feature integration. 
Variations of the model that only leverage one feature exhibit higher variance and reduced accuracy, particularly at moderate to high missing rates. In contrast, the full \texttt{TANDEM} framework with multi-head attention and gating consistently yields narrower box plots, reflecting lower variance and higher median accuracy. 
%

\section{Real-world data analysis}\label{sec:Real_world}
To evaluate the effectiveness of our proposed \texttt{TANDEM} framework in a real-world scenario, we conducted an additional study using a medical dataset, `PhysioNet Sepsis'. 
The 2019 PhysioNet / Computing in Cardiology (CinC) challenge on Sepsis prediction serves as the foundation for time-series classification experiments~\citep{reyna_early_2020}. 
The dataset used in these experiments contains 40,335 cases of patients admitted to intensive care units, with 34 time-dependent variables such as heart rate, oxygen saturation, and body temperature. The primary objective is to classify whether each patient has sepsis or not according to the sepsis-3 definition.

\paragraph{Experimental Setup. } The PhysioNet dataset is an irregular time series dataset, as only 10\% of the values are sampled with their respective timestamps for each patient. To address this irregularity, two types of time-series classification are performed: (OI) classification using observation intensity, and (no OI) classification without using observation intensity.
In medical context, observation intensity is a measure of the degree of illness, and when incorporated, an index number is appended to each value in the time series. 

\subsection{Results analysis}
Given the data's imbalanced nature, we evaluate performance using the Area Under the Receiver Operating Characteristic (AUROC) score.
We used the reported performance from \citet{jhin_attentive_2024} and \citet{oh_stable_2024} to compare with our method. 
In Table~\ref{tab:sepsis}, our \texttt{TANDEM} framework demonstrated robust performance in both OI and no OI settings. The incorporation of observation intensity generally improved the classification performance, indicating the importance of explicit features. 
These results underscore the potential of our \texttt{TANDEM} framework in real-world medical applications. 
\begin{table}[htb]
\scriptsize\centering\captionsetup{skip=5pt}
\caption{AUROC on PhysioNet Sepsis}\label{tab:sepsis}
\begin{tabular}{@{}lcc@{}}
\toprule
\multicolumn{1}{c}{\multirow{2.5}{*}{\textbf{Method}}} & \multicolumn{2}{c}{\textbf{Test AUROC}}         \\ \cmidrule(l){2-3} 
\multicolumn{1}{c}{}                               & \textbf{OI}            & \textbf{No OI}         \\ \midrule
GRU-$\Delta t$                                     & 0.878\std{0.006}          & 0.840\std{0.007}          \\
GRU-D                                              & 0.871\std{0.022}          & 0.850\std{0.013}          \\
GRU-ODE                                            & 0.852\std{0.010}          & 0.771\std{0.024}          \\
ODE-RNN                                            & 0.874\std{0.016}          & 0.833\std{0.020}          \\
Latent-ODE                                         & 0.787\std{0.011}          & 0.495\std{0.002}          \\
ACE-NODE                                           & 0.804\std{0.010}          & 0.514\std{0.003}          \\
Neural CDE                                         & 0.880\std{0.006}          & 0.776\std{0.009}          \\
ANCDE                                              & 0.900\std{0.002}          & 0.823\std{0.003}          \\
EXIT                                               & {\ul 0.913\std{0.002}}    & 0.836\std{0.003}          \\
Neural LSDE                                        & 0.909\std{0.004}          & 0.879\std{0.008}          \\
Neural LNSDE                                       & 0.911\std{0.002}          & 0.881\std{0.002}          \\
Neural GSDE                                        & 0.909\std{0.001}          & {\ul 0.884\std{0.002}}    \\ \midrule
\textbf{Neural ODE (with \texttt{TANDEM})}         & \textbf{0.916\std{0.005}} & 0.877\std{0.008}          \\
\textbf{Neural CDE (with \texttt{TANDEM})}         & 0.909\std{0.006}          & \textbf{0.893\std{0.003}} \\
\textbf{Neural SDE (with \texttt{TANDEM})}         & {\ul 0.913\std{0.005}}    & 0.873\std{0.004}          \\ \bottomrule
\end{tabular}
\end{table}

Moreover, our findings support the hypothesis that missing data patterns can contribute to a deeper understanding of time series. 
While attention applied to learned latent features shows marginal improvements, incorporating attention on raw observation or on the control path leads to more substantial performance gains. This indicates that missingness patterns encode predictive signals, offering useful context beyond the observed values themselves.

\begin{table}[htb]
\scriptsize\centering\captionsetup{skip=5pt}
\caption{Average gate activations on PhysioNet Sepsis}\label{tab:gating_sepsis}
\begin{tabular}{@{}cccc@{}}
\toprule
\multicolumn{2}{c}{\textbf{Model Component}}                                                                                             & \textbf{OI}      & \textbf{No OI}   \\ \midrule
\multirow{3}{*}{\textbf{\begin{tabular}[c]{@{}c@{}}Neural   ODE \\      (with $\texttt{TANDEM}$)\end{tabular}}} & $\sigma_{\tilde{\vx}}$ & 0.600\std{0.548} & 0.400\std{0.548} \\
                                                                                                                & $\sigma_{\bm{X}}$      & 0.600\std{0.548} & 1.000\std{0.000} \\
                                                                                                                & $\sigma_{\vz}$         & 1.000\std{0.000} & 1.000\std{0.000} \\ \midrule
\multirow{3}{*}{\textbf{\begin{tabular}[c]{@{}c@{}}Neural CDE \\      (with $\texttt{TANDEM}$)\end{tabular}}}   & $\sigma_{\tilde{\vx}}$ & 0.400\std{0.548} & 0.800\std{0.447} \\
                                                                                                                & $\sigma_{\bm{X}}$      & 0.000\std{0.000} & 0.400\std{0.548} \\
                                                                                                                & $\sigma_{\vz}$         & 1.000\std{0.000} & 1.000\std{0.000} \\ \midrule
\multirow{3}{*}{\textbf{\begin{tabular}[c]{@{}c@{}}Neural SDE \\      (with $\texttt{TANDEM}$)\end{tabular}}}   & $\sigma_{\tilde{\vx}}$ & 0.800\std{0.447} & 0.200\std{0.447} \\
                                                                                                                & $\sigma_{\bm{X}}$      & 1.000\std{0.000} & 1.000\std{0.000} \\
                                                                                                                & $\sigma_{\vz}$         & 1.000\std{0.000} & 1.000\std{0.000} \\ \bottomrule
\end{tabular}
\end{table}

Table~\ref{tab:gating_sepsis} summarizes the average gate activations under two conditions. 
The latent gate $\sigma_{\vz}$ maintains consistently high values, indicating the importance of learned continuous-time dynamics. The gates $\sigma_{\tilde{\vx}}$ and $\sigma_{\bm{X}}$ show moderate values that adapt based on the presence of observation intensity features, while preserving the model's reliance on its continuous-time backbone.
Overall, the proposed \texttt{Gumbel-Sigmoid} gating behavior confirms $\texttt{TANDEM}$'s flexibility in utilizing the most informative features across different data scenarios.

\subsection{Ablation study}
To assess the efficacy of our approach, we conducted a comprehensive ablation study, with results presented in Table~\ref{tab:sepsis_ablation}. This analysis reveals that the \texttt{TANDEM} framework consistently improves the performance of NDE-based models compared to their respective baselines. These enhancements highlight the effectiveness of our proposed attention-guided mechanism in addressing the complexities of time series data.
\begin{table}[htb]
\scriptsize\centering\captionsetup{skip=5pt}
\caption{Ablation study on PhysioNet Sepsis}\label{tab:sepsis_ablation}
\begin{tabular}{@{}clcc@{}}
\toprule
\multirow{2.5}{*}{\textbf{Baseline}} & \multicolumn{1}{c}{\multirow{2.5}{*}{\textbf{Component}}} & \multicolumn{2}{c}{\textbf{Test AUROC}} \\ \cmidrule(l){3-4} 
                                   & \multicolumn{1}{c}{}                                    & \textbf{OI}        & \textbf{No OI}     \\ \midrule
\multirow{7}{*}{\textbf{Neural ODE}} & \texttt{TANDEM}                                         & 0.916\std{0.005} & 0.877\std{0.008}     \\
                            & \texttt{TANDEM} (soft)                                  & 0.898\std{0.007} & 0.860\std{0.012}     \\
                            & $[\Phi_{z(t)}$,   $\Phi_{\tilde{x}(t)}$, $\Phi_{X(t)}]$ & 0.912\std{0.007} & 0.864\std{0.009}     \\
                            & $[\Phi_{z(t)}$, $\Phi_{X(t)}]$                          & 0.911\std{0.004} & 0.868\std{0.020}     \\
                            & $[\Phi_{z(t)}$,   $\Phi_{\tilde{x}(t)}]$                & 0.909\std{0.004} & 0.848\std{0.005}     \\
                            & $\Phi_{z(t)}$ only                                      & 0.849\std{0.012} & 0.846\std{0.005}     \\
                            & w/o attention                                           & 0.844\std{0.008} & 0.851\std{0.006}     \\ \midrule
\multirow{7}{*}{\textbf{Neural CDE}} & \texttt{TANDEM}                                         & 0.909\std{0.006} & 0.893\std{0.003}     \\
                            & \texttt{TANDEM} (soft)                                  & 0.896\std{0.005} & 0.885\std{0.005}     \\
                            & $[\Phi_{z(t)}$,   $\Phi_{\tilde{x}(t)}$, $\Phi_{X(t)}]$ & 0.892\std{0.003} & 0.887\std{0.004}     \\
                            & $[\Phi_{z(t)}$, $\Phi_{X(t)}]$                          & 0.902\std{0.005} & 0.885\std{0.003}     \\
                            & $[\Phi_{z(t)}$,   $\Phi_{\tilde{x}(t)}]$                & 0.904\std{0.003} & 0.889\std{0.006}     \\
                            & $\Phi_{z(t)}$ only                                      & 0.907\std{0.008} & 0.886\std{0.005}     \\
                            & w/o attention                                           & 0.888\std{0.002} & 0.864\std{0.005}     \\ \midrule
\multirow{7}{*}{\textbf{Neural SDE}} & \texttt{TANDEM}                                         & 0.913\std{0.005} & 0.873\std{0.004}     \\
                            & \texttt{TANDEM} (soft)                                  & 0.909\std{0.003} & 0.853\std{0.013}     \\
                            & $[\Phi_{z(t)}$,   $\Phi_{\tilde{x}(t)}$, $\Phi_{X(t)}]$ & 0.911\std{0.007} & 0.861\std{0.006}     \\
                            & $[\Phi_{z(t)}$, $\Phi_{X(t)}]$                          & 0.903\std{0.006} & 0.872\std{0.006}     \\
                            & $[\Phi_{z(t)}$,   $\Phi_{\tilde{x}(t)}]$                & 0.902\std{0.009} & 0.847\std{0.005}     \\
                            & $\Phi_{z(t)}$ only                                      & 0.846\std{0.008} & 0.846\std{0.006}     \\
                            & w/o attention                                           & 0.850\std{0.004} & 0.848\std{0.003}     \\ \bottomrule
\end{tabular}
\end{table}

The improved AUROC scores achieved by our \texttt{TANDEM} framework on the PhysioNet Sepsis dataset suggest that it can effectively aid in early detection of sepsis, potentially leading to timely interventions and improved patient outcomes. Our method can be valuable in healthcare and clinical settings where data collection is often incomplete or irregular.

\section{Conclusion}\label{sec:Conclusion}

We introduced a novel attention-guided NDE framework for time series classification in the presence of missing values. Our approach combines the strengths of attention mechanisms and neural differential equations to capture the underlying dynamics of incomplete time series data and improve classification performance. 
Integrating attention mechanisms with NDEs enables a more robust and accurate classification of time series containing missing values.

While our framework demonstrates significant improvements, it also presents certain limitations. The computational complexity of training NDEs with attention mechanisms can be high, especially for large-scale datasets. Future research could explore more efficient training algorithms or approximation methods to address scalability. Additionally, extending the framework to other tasks, and investigating its performance under different missingness mechanisms, are promising directions for further study.

\begin{acks}
This research was supported by 
the Basic Science Research Program through the National Research Foundation of Korea (NRF) funded by the Ministry of Education (RS-2024-00407852); 
the Korea Health Technology R\&D Project through the Korea Health Industry Development Institute (KHIDI), funded by the Ministry of Health and Welfare, Republic of Korea (HI19C1095); 
the Ministry of Trade, Industry and Energy (MOTIE) and Korea Institute for Advancement of Technology (KIAT) through the International Cooperative R\&D program (No.P0025828); 
Institute of Information \& communications Technology Planning \& Evaluation (IITP) grant funded by the Korea government (MSIT)
(No. RS-2020-II201336, Artificial Intelligence Graduate School Program (UNIST)); 
and the National Research Foundation of Korea (NRF) grant funded by the Korea government (MSIT) (RS-2025-00563597).
\end{acks}

\section*{GenAI Usage Disclosure}
We used Generative AI tools (e.g., large language models) solely for language editing and rephrasing. No AI-generated content was used to propose novel ideas, or conduct experiments. All intellectual contributions are solely those of the authors.

\bibliographystyle{ACM-Reference-Format}
\balance
\bibliography{references}


\end{document}